%% file: iclr2026_conference.tex
\documentclass{article} 
\usepackage{iclr2026_conference,times}
\iclrfinalcopy
\input{math_commands.tex}

\usepackage{tabularx}
\usepackage{hyperref}
\usepackage{url}
\usepackage{graphicx}
\usepackage{subcaption}
\usepackage{enumitem}
\usepackage{multirow}
\usepackage{multicol}
\usepackage{graphicx}
\usepackage{tabularx}
\usepackage{colortbl}
\usepackage{amsmath} 
\usepackage{wrapfig}
\usepackage{float}

\definecolor{lightblue}{HTML}{E0ECFF}
\definecolor{lightred}{HTML}{FFD4DA}
\definecolor{platinum}{HTML}{E5E4E2}
\definecolor{rowyellow}{HTML}{FFF9E6} 
\definecolor{rowblue}{HTML}{E8F4FF} 
\definecolor{rowgreen}{HTML}{EBF7E3} 
\definecolor{ourgreen}{HTML}{F0FFF2}
\definecolor{ourred}{HTML}{FFF5F5}

\definecolor{promptgrey}{HTML}{F2F3F4}

\definecolor{mylightblue}{RGB}{30, 144, 255} 
 \usepackage{float}
\definecolor{mygreen}{RGB}{75, 225, 75}
\hypersetup{
    colorlinks=true,       
    citecolor=mylightblue,
}
 
\usepackage[table]{xcolor} 
\usepackage{booktabs}      
\usepackage{multirow}      
\usepackage{arydshln}      
\usepackage{amsmath}       
\usepackage{wrapfig}
\definecolor{citeblue}{HTML}{2774AE} 
\definecolor{rowgray}{gray}{0.92}    
\definecolor{textgray}{gray}{0.65} 

\title{VideoRAE: Taming Video Foundation Models for Generative Modeling via Representation Autoencoders}

\author{
Zhihao Xie$^{1,2,}$\thanks{Equal contribution.} \qquad 
Junfeng Wu$^{2,}$\footnotemark[1] \qquad 
Xinting Hu$^{4}$ \qquad
Junchao Huang$^{1,3}$ \qquad
Li Jiang$^{1,3,}$\thanks{Corresponding author.} \\[0.15cm]
$^1$ The Chinese University of Hong Kong, Shenzhen \\
$^2$ Huazhong University of Science and Technology  \quad $^3$ Shenzhen Loop Area Institute \\
$^4$ University of Science and Technology of China \\[0.15cm]
}

\begin{document}

\maketitle

\vspace{-20pt} 
\begin{center}
    \includegraphics[width=1.0\textwidth]{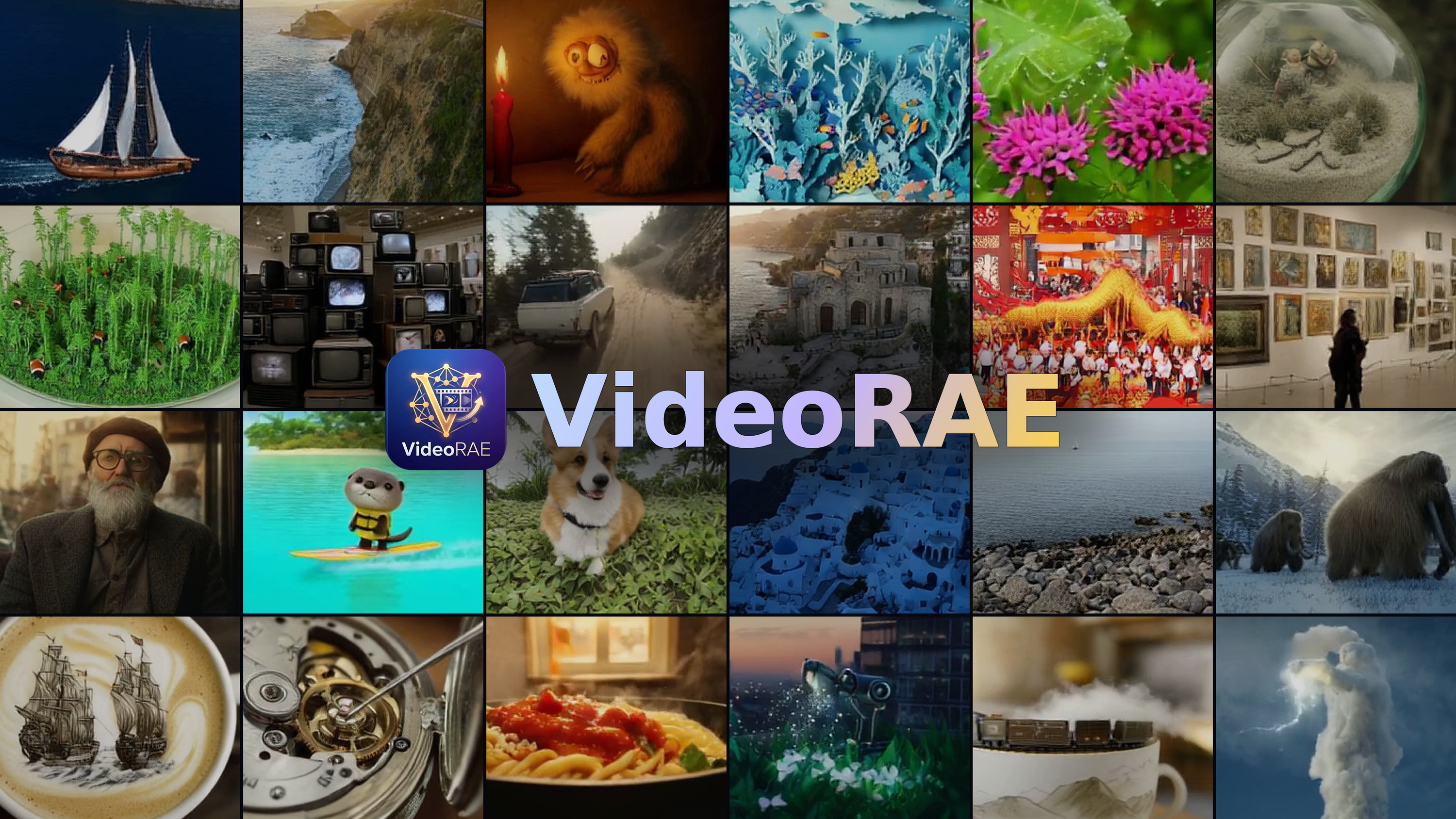}
    \captionof{figure}{Text-to-video samples generated by an 11B OpenSora2 model adapted to VideoRAE-3D.}
    \label{fig:videoraet2v}
\end{center}
\vspace{10pt}

    \begin{abstract}

    Video generation models typically rely on 3D-VAEs trained for pixel-level reconstruction, whose latent spaces may underrepresent semantic structure. We introduce VideoRAE, a representation autoencoder that converts features from a frozen video foundation model into compact, reconstruction-capable latents for video generation. A lightweight 1D self-attention projector compresses multi-scale hierarchical features, producing continuous latents for Diffusion Transformers and discrete tokens for autoregressive models through multi-codebook high-dimensional quantization. During decoding, a local–global representation alignment objective transfers semantic structure from the frozen encoder and removes the need for KL regularization. Comprehensive experiments show that VideoRAE achieves strong reconstruction in both continuous and discrete regimes. On UCF-101, autoregressive and diffusion generators built on VideoRAE achieve class-conditional gFVD scores of 40 and 93, respectively, while converging approximately \(5\times\) faster than autoencoder baselines. In controlled 2B-parameter text-to-video experiments, replacing LTX-VAE with VideoRAE accelerates convergence and consistently improves VBench performance. These results establish frozen video foundation representations as compact, versatile, and generation-friendly video latents. Code and models are \href{https://zhxie0117.github.io/VideoRAE/}{released}.

\end{abstract}

\section{Introduction}
The field of generative modeling has witnessed a paradigm shift, transitioning from pixel-space generation \citep{van2016pixel,goodfellow2014gan} to latent-space generation \citep{ldm,llamagen,VAE,van2017vqvae}. For video synthesis, both Variational Autoencoders (VAEs) \citep{cheng2025leanvae,wan2025wan,yang2025cogvideox} and discrete tokenizers \citep{wang2024omnitokenizer,hong2022cogvideo} 
typically employ 
3D visual encoders to compress raw video clips into compact spatio-temporal tokens. 
This approach effectively reduces computational complexity. However, existing video tokenizers are still predominantly optimized for pixel-level reconstruction, which can underemphasize semantic and long-range spatio-temporal structure.
Driven by pixel-level Mean Squared Error and adversarial losses, the latent spaces of these visual tokenizers are biased towards optimizing local appearance, rather than capturing high-level semantics or physical structures. Consequently, downstream generative models are burdened with the heavy lifting of learning complex spatio-temporal dynamics and global semantics, 
increasing both the difficulty and data requirements of training.

Concurrently, visual representation learning has made substantial progress.
Video Foundation Models, represented by V-JEPA 2 \citep{assran2025vjepa2} and VideoMAEv2 \citep{wang2023videomae}, have shown strong capabilities in extracting structured and predictive spatio-temporal features. In the image domain, recent representation autoencoders such as RAE \citep{zheng2025rae} and VFMTok \citep{zheng2025vfmtok} suggest that pretrained visual representations can be converted into decodable and generation-friendly latent spaces. However, extending this idea to video is nontrivial. Videos contain richer temporal dynamics, stronger redundancy, and stricter compression requirements than images, making it unclear whether frozen VFM features can be directly used as compact 
and generation-compatible 
video latents. 
Existing tokenizer studies for video generation
\citep{guo2025dera,zhang2025videorepa,xiong2026evatok} have mainly incorporated semantic alignment as auxiliary supervision within conventional pixel-driven VAE frameworks, 
while largely retaining their original encoder architectures and latent construction mechanisms.
Therefore, 
whether frozen VFM representations can be transformed into compact
latent spaces that support both high-fidelity reconstruction and downstream video
generation remains underexplored.

To address this question, we introduce VideoRAE, a unified video autoencoding
framework that uses a frozen VFM as its encoder. VideoRAE aggregates
hierarchical VFM features to capture both coarse semantics and fine-grained
spatio-temporal dynamics, and compresses these redundant representations with
a lightweight 1D self-attention projector to meet the stringent compression
requirements of video generation. During decoding, we further apply a local-and-global representation alignment objective with the frozen VFM teacher, which
encourages the decoder to preserve the semantic structure of the foundation
representation and enables stable training without KL regularization. The same framework supports both major latent generative paradigms: continuous
latents for diffusion transformers, and discrete tokens for autoregressive models
via a multi-codebook high-dimensional quantization strategy that preserves the
capacity of VFM representations while making them autoregressively tractable.

\begin{figure*}[t]
    \vspace{-5mm}
    \centering
    \includegraphics[width=1.0\linewidth]{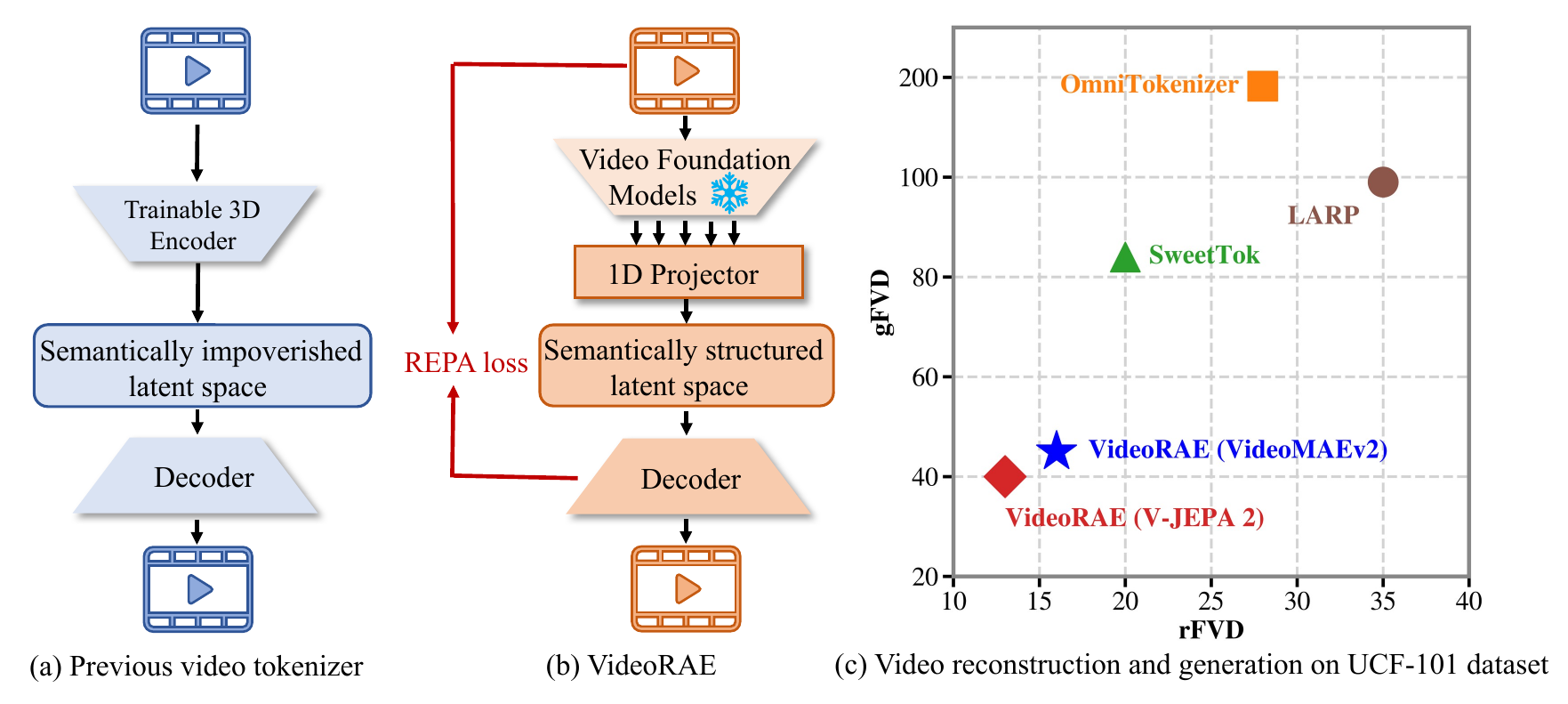}
    \vspace{-5mm}
    \caption{Conceptual comparison between traditional video tokenizers and VideoRAE. Traditional 3D-VAEs are trained from scratch and driven solely by pixel-level MSE and adversarial losses. In contrast, VideoRAE directly utilizes a frozen video foundation model as its core feature extractor. VideoRAE provides a unified and semantically rich latent space that perfectly supports both continuous and discrete generative paradigms, achieving high-fidelity reconstruction and high-quality generation.}
    \label{fig:intro}
    \vspace{-13pt}
\end{figure*}

We comprehensively evaluate VideoRAE across both continuous and discrete latent
regimes, covering video reconstruction and downstream generative modeling. On
UCF-101 and TokenBench, VideoRAE achieves state-of-the-art discrete
reconstruction and highly competitive continuous reconstruction, demonstrating
that frozen VFM representations can remain reconstruction-capable after strong
compression. For class-to-video generation on UCF-101, its discrete AR and
continuous DiT variants achieve state-of-the-art gFVDs of 40 and 93,
respectively. Notably, the AR variant outperforms prior discrete tokenizers while
using fewer tokens and a smaller generator, and both AR and DiT generators
converge approximately  $5\times$ faster than their respective autoencoder baselines. We
further conduct a controlled 2B-scale text-to-video replacement study, where
VideoRAE leads to faster convergence than the LTX-VAE baseline under the same
training framework and comparable settings. Together, these results demonstrate
that the advantages of VideoRAE hold across latent formulations, generative
paradigms, and task complexity.

Our main contributions are summarized as follows:
\vspace{-0.5em} 
\begin{itemize}
    \setlength{\itemsep}{2pt}  
    \setlength{\parsep}{0pt}   
    \setlength{\parskip}{0pt}

    \item We diverge from the conventional paradigm that restricts frozen Video Foundation Models (VFMs) primarily to understanding tasks. Instead, we systematically demonstrate for the first time that frozen VFMs can be unleashed to serve as robust, directly applicable autoencoder foundations for high-fidelity video generation.


    \item We propose VideoRAE, featuring a semantic-driven 1D projector and a local-and-global representation alignment objective tailored to frozen VFM features. It unifies continuous KL-free latents and discrete multi-codebook latents within a single framework.

    \item Extensive experiments demonstrate that VideoRAE not only matches traditional VAEs in reconstruction metrics but also provides a structured semantic latent space that significantly accelerates the convergence of downstream generative models and improves generation quality. Our work confirms the immense feasibility and potential of visual foundation model representations in video generative tasks.
\end{itemize}
\vspace{-0.5em}

\section{Related Works}

\subsection{Video Autoencoders and Tokenizers}
\label{sec:related_video_vae}
Video variational autoencoders (VAEs) and tokenizers are crucial components for compressing high-dimensional video data into compact latent spaces, which is essential for realizing efficient generative modeling. In the realm of continuous spaces, VFRTok \citep{zhong2026vfrtok} attempts to enhance model robustness through variable frame rate training, VidTwin \citep{wang2025vidtwin} improves reconstruction quality by decoupling structure and dynamics, and LeanVAE \citep{cheng2025leanvae} boosts training efficiency via lightweight network design and wavelet transforms. In the domain of discrete tokenization, LARP \citep{wang2024larp} pioneers a 1D tokenizer architecture and elevates generation quality through an autoregressive prior, whereas SweetTok \citep{tan2025sweettok} mitigates quantization loss by utilizing a dual-stream spatio-temporal codebook. Despite their success, these models are typically trained entirely from scratch, relying heavily on a combination of pixel-level reconstruction and adversarial losses. Unlike these pixel-driven approaches, VideoRAE explores a semantic-driven compression paradigm by directly leveraging frozen VFMs, thereby providing a highly structured latent space that alleviates the learning burden on downstream generators. 

\subsection{Visual Foundation Models}
\label{sec:related_vfm}
Visual Foundation Models (VFMs) \citep{oquab2023dinov2,simeoni2025dinov3,assran2025vjepa2,murlabadia2026vjepa2_1,wang2023videomae,wang2022internvideo} aim to learn general, transferable representations from large-scale, diverse data. In the image domain, self-supervised learning and language-supervised pre-training on image-text pairs have endowed VFMs with rich and semantically grounded representations. Extending to the video domain, the training of video foundation models is predominantly based on scalable self-supervised learning, leveraging the inherent spatio-temporal structures of videos. Notably, VideoMAEv2 \citep{wang2023videomae} scales masked video autoencoding to billion-level parameters via a dual-masking strategy, forcing the model to develop robust spatio-temporal reasoning capabilities from extremely sparse inputs. Similarly, V-JEPA 2 \citep{assran2025vjepa2} employs a joint-embedding predictive architecture. By having the target encoder predict the latent features of masked regions, V-JEPA 2 \citep{assran2025vjepa2} deliberately discards low-level visual noise to primarily focus on high-level semantics and physical motion logic. 
While these VFMs exhibit exceptional understanding capabilities, their highly abstract nature has historically deterred their direct application in high-fidelity pixel reconstruction,
which leaves open whether such abstract video representations can be directly
used for high-fidelity video generation.

\subsection{Representation Learning and Generation}
\label{sec:related_rep_gen}
Traditionally, representations learned for visual understanding \citep{tschannen2025siglip,wang2023videomae,simeoni2025dinov3,oquab2023dinov2} and those optimized for generation \citep{llamagen,esser2021tamingvqgan,sdxl,yao2025vavae,VAE,van2017vqvae,xia2026reworld} have evolved along two distinct trajectories. Understanding-oriented representations focus on extracting high-level semantics while discarding low-level noise, whereas generation-oriented latents prioritize capturing high-fidelity textures. However, with the widespread adoption of representation alignment, recent advancements are rapidly blurring this boundary. In the image domain, works such as RAE \citep{zheng2025rae} and VFMTok \citep{zheng2025vfmtok} demonstrate that features extracted from frozen visual foundation models can serve as highly expressive latent spaces for image generation. Furthermore, MAETok \citep{chen2025masked} bridges self-supervised features with discrete token-based generation, while works like UniTok \citep{ma2025unitok} attempt to unify generation and understanding tasks. These pioneering efforts share a common insight: operating within a highly semantic latent space can significantly improve both generation quality and training efficiency. Despite these advances, in the specific realm of video autoencoding, semantic features are still typically relegated to mere auxiliary supervision signals. In this paper, we introduce VideoRAE to demonstrate that features from video foundation models can inherently serve as exceptional representations for visual generation.

\section{Method: VideoRAE}

The core philosophy of VideoRAE is to abandon the traditional pixel-driven paradigm of training video tokenizers from scratch. Instead, we propose a semantic-driven compression framework built directly upon frozen Video Foundation Models (VFMs). Given an input video (Sec.~\ref{sec:encoding}), we first utilize a pre-trained and frozen VFM (e.g., V-JEPA 2) to extract hierarchical semantic features spanning from shallow to deep layers. Next, we design a lightweight 1D projector to perform spatio-temporal fusion and downsampling on these features, generating highly compact tokens. Subsequently (Sec.~\ref{sec:latent}), depending on the requirements of downstream generative tasks, these tokens are seamlessly mapped into either a continuous representation (for Diffusion Transformers) or a discrete representation (for autoregressive models). Finally (Sec.~\ref{sec:repa}), a decoder is tasked with reconstructing the original pixels from these latent representations. During this process, we introduce a Representation Alignment (REPA) module to guide the decoder's learning. By explicitly aligning semantics, we intrinsically maintain a high-quality topological structure in the latent space, which completely replaces the traditional KL-divergence constraint.

\subsection{Semantic-Driven Encoder and 1D Projector}
\label{sec:encoding}

Given an input video clip $X \in \mathbb{R}^{B \times 3 \times T \times H \times W}$, we leverage an off-the-shelf, frozen VFM as the primary feature extractor. To simultaneously capture low-level temporal dynamics and high-level abstract semantics, we extract hierarchical intermediate features from shallow to deep layers of the VFM and explicitly fuse them via element-wise summation:
\begin{equation}
    F_{\text{VFM}} = \sum_{l \in \mathcal{S}} f_l(X), \quad F_{\text{VFM}} \in \mathbb{R}^{B \times N_{\text{vfm}} \times D_{\text{token}}}
\end{equation}
where $\mathcal{S}$ represents the set of selected layers, and $f_l(\cdot)$ denotes the mapping function of the $l$-th VFM layer. Here, $N_{\text{vfm}}$ represents the total number of flattened 3D spatio-temporal tokens output by the VFM, and $D_{\text{token}}$ is the embedding dimension.

Although $F_{\text{VFM}}$ encapsulates highly expressive hierarchical semantics, its extensive sequence length ($N_{\text{vfm}}$) and inherent spatio-temporal redundancy pose significant computational bottlenecks for downstream generative modeling. To address this, we introduce a lightweight projector, denoted as $\mathcal{E}_{\text{proj}}$, to dynamically condense the long spatio-temporal sequence into a highly compact sequence of base tokens:
\begin{equation}
    Z_{\text{base}} = \mathcal{E}_{\text{proj}}(F_{\text{VFM}}), \quad Z_{\text{base}} \in \mathbb{R}^{B \times N_{\text{latent}} \times D_{\text{token}}}
\end{equation}
where $N_{\text{latent}}$ is the length of the compressed latent sequence. By default, $\mathcal{E}_{\text{proj}}$ acts as a 1D projector that treats the VFM features as a flattened 1D sequence. Specifically, it concatenates a fixed number ($N_{\text{latent}}$) of learnable tokens to the input and employs self-attention mechanisms to aggregate global spatio-temporal information into these tokens. The updated learnable tokens are then extracted to form the compressed latent sequence $Z_{\text{base}}$. Additionally, for the continuous version of VideoRAE, to demonstrate the generality of our method, we also explore a 3D projector variant. This variant explicitly preserves the spatio-temporal structure by applying a 3D convolutional layer for local compression, followed by a few simple self-attention layers to capture global context. Unless otherwise specified, the 1D projector serves as the default configuration throughout the remainder of this paper.

\begin{figure*}[t]
    \centering
    \includegraphics[width=1.0\linewidth]{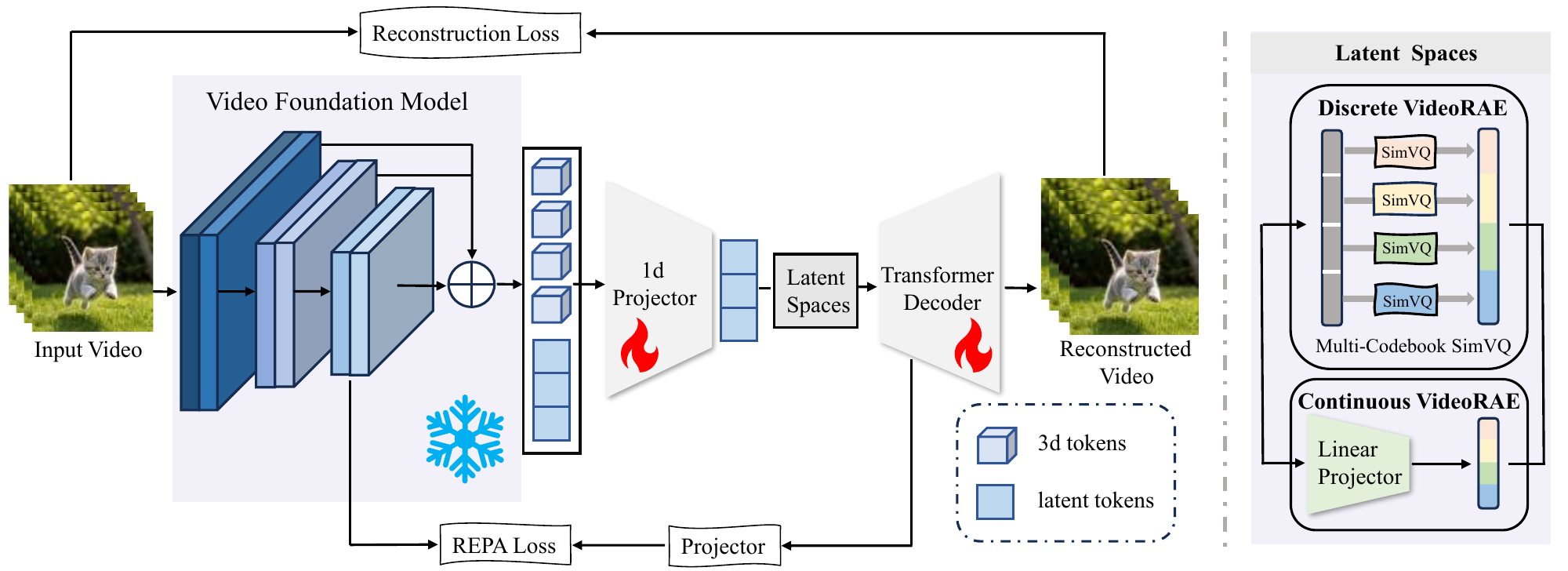}
    \vspace{-5mm}
    \caption{\textbf{Overall architecture of VideoRAE.} Given an input video, multi-scale hierarchical features are first extracted from a frozen VFM and fused, followed by a 1D self-attention projector that dynamically compresses them into compact base tokens. These tokens are then formatted into either a continuous latent space via linear projection for Diffusion Transformers, or a discrete latent space via Multi-Codebook SimVQ for Autoregressive models. Finally, during decoding, the Representation Alignment module explicitly aligns intermediate decoder features with the VFM teacher at both local and global scales, intrinsically regularizing the semantic structure of the latent manifold.}
    \label{fig:pipeline}
\end{figure*}

\subsection{Versatile Generative Latent Spaces}
\label{sec:latent}
To cater to the distinct preferences of various generative architectures, VideoRAE is designed to format the base tokens into specific latent spaces.

\vspace{1mm}
\textbf{Continuous VideoRAE for Diffusion Models.} 
For generative architectures like Diffusion Transformers (DiTs), a smooth and continuous latent space is preferred. Traditional continuous 3D-VAEs heavily rely on the KL-divergence penalty to explicitly regularize the latent space into a standard Gaussian distribution. However, this rigid constraint often restricts the expressive capacity of the latents, leading to the over-smoothing of the visual details and causing severe optimization conflicts with the pixel-reconstruction objective. In contrast, VideoRAE completely bypasses the need for the KL-divergence penalty. By inheriting the highly structured semantic topology from the powerful VFM encoder, and implicitly regularizing the representation through our subsequent Representation Alignment (REPA) module, the latent space is intrinsically maintained on a well-behaved semantic manifold. Consequently, without imposing any artificial distribution assumptions, a simple linear layer effectively projects $Z_{\text{base}}$ into a latent space, yielding a highly expressive and DiT-friendly continuous latent $Z_{\text{cont}}$.

\vspace{1mm}
\textbf{Discrete VideoRAE for Autoregressive Models.} Autoregressive generative models rely on discrete token sequences. To break the quantization bottleneck and maximally preserve the rich semantics inherited from the VFM, a straightforward solution is to increase both the codebook size and the latent dimensionality. However, existing studies indicate that simply expanding the codebook yields diminishing returns. Moreover, under strong semantic constraints, simple vector quantization easily causes a large portion of codewords to remain underutilized during training, ultimately triggering severe codebook collapse.

To address this, we propose the \textbf{Multi-Codebook SimVQ} strategy, operating on a divide-and-conquer principle. Specifically, the continuous latent vector $Z_{\text{base}}$ is first evenly partitioned along the channel dimension into $K$ sub-vectors: $Z_{\text{base}} = [Z_1, Z_2, \dots, Z_K]$, where $Z_k \in \mathbb{R}^{B \times N_{\text{latent}} \times (D_{\text{token}}/K)}$. To stabilize the quantization of high-dimensional semantic features, we integrate the SimVQ paradigm into each sub-codebook. Unlike standard VQ that directly optimizes the codebook embeddings, SimVQ initializes a frozen base codebook $E_k$ and maps it to the target latent space via a learnable Multi-Layer Perceptron (MLP). Thus, the effective sub-codebook $\mathcal{C}_k$ is generated dynamically, and only the MLP parameters are updated during training. For each chunk $Z_k$, its quantized representation $\hat{Z}_k$ is obtained by searching for the nearest neighbor in the mapped sub-codebook $\mathcal{C}_k$:
\begin{equation}
    \mathcal{C}_k = \text{MLP}_k (E_k), \quad \hat{Z}_k = \underset{c \in \mathcal{C}_k}{\arg\min} \| Z_k - c \|^2_2
\end{equation}
The final discrete latent $Z_{\text{disc}}$ is constructed by concatenating these quantized chunks: $Z_{\text{disc}} = [\hat{Z}_1, \hat{Z}_2, \dots, \hat{Z}_K]$.

Compared to conventional methods, the multi-codebook strategy effectively scales up the vocabulary size while keeping individual codebooks within an easily optimizable range. More importantly, preserving a high-dimensional latent space is crucial for maintaining the semantic integrity of the VFM. The semantic features extracted by VFMs are inherently complex and highly entangled. Forcing them into a low-dimensional bottleneck inevitably discards critical context and fine-grained physical dynamics. By maintaining a high overall dimensionality through the multi-codebook mechanism, the quantized discrete representations can comfortably accommodate the rich information provided by the VFM. This design not only substantially prevents model collapse but also provides ample semantic guidance for the subsequent decoding process under REPA constraints.

\begin{figure*}[!htb]
    \centering
    \includegraphics[width=1.0\linewidth]{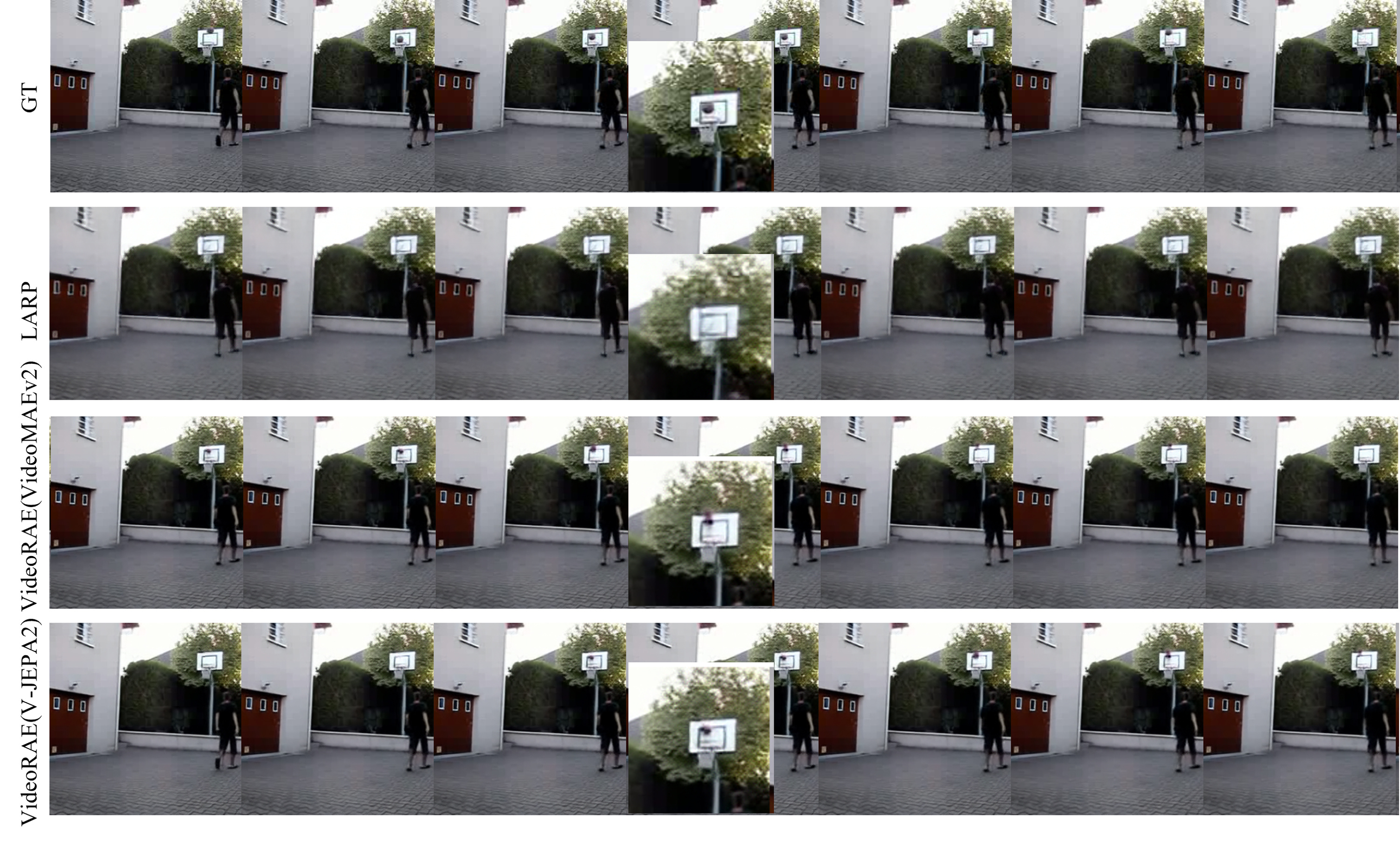}
    \vspace{-5mm}
    \caption{Visualization comparison of the Discrete-token reconstruction on the UCF-101 dataset.}
    \label{fig:vis2}
    \vspace{-13pt}
\end{figure*}

\subsection{Decoder and Representation Alignment}
\label{sec:repa}
The decoder $\mathcal{D}_{\text{sem}}$ is responsible for reconstructing original video pixels from the highly compressed latent representations. Specifically, $\mathcal{D}_{\text{sem}}$ is designed as a Transformer-based architecture. It accepts the sequence of 1D latent tokens and concatenates them with $1024$ learnable 3D tokens. After processing these concatenated tokens through stacked self-attention blocks, the learnable 3D tokens are separated and mapped back into 3D feature maps via a depatchify operation for subsequent pixel reconstruction.

Beyond traditional reconstruction losses, we propose the Representation Alignment (REPA) objective. Let $H_{\text{stu}}$ denote the shallow 3D intermediate features of the decoder (i.e., the spatial features derived shortly after the depatchify operation). We project and spatially interpolate $H_{\text{stu}}$ across a 3D grid to align with the final layer features of the VFM, $F_{\text{target}}$, yielding the aligned feature $H_{\text{align}}$.

The REPA loss enforces semantic consistency at both local and global scales:
\begin{equation}
    \mathcal{L}_{\text{local}} = - \frac{1}{N_{\text{vfm}}} \sum_{i=1}^{N_{\text{vfm}}} \frac{H_{\text{align}}^{(i)} \cdot F_{\text{target}}^{(i)}}{\|H_{\text{align}}^{(i)}\| \|F_{\text{target}}^{(i)}\|}
\end{equation}
\begin{equation}
    \mathcal{L}_{\text{global}} = - \frac{\text{Pool}(H_{\text{align}}) \cdot \text{Pool}(F_{\text{target}})}{\|\text{Pool}(H_{\text{align}})\| \|\text{Pool}(F_{\text{target}})\|}
\end{equation}
\begin{equation}
    \mathcal{L}_{\text{REPA}} = \mathcal{L}_{\text{local}} + \lambda \mathcal{L}_{\text{global}}
\end{equation}
where $\lambda$ is the weight for the global loss. Crucially, this semantic alignment implicitly regularizes the geometry of the latent space, forcing the representations to reside on a well-behaved semantic manifold. It forces the latent features to adhere to the high-quality semantic topological structure of the foundation model. This structural preservation is the fundamental reason why we can 
discard the KL-divergence loss used in traditional 3D-VAEs in our VideoRAE.

\subsection{Training Objectives}
The end-to-end training of VideoRAE integrates pixel-level reconstruction, adversarial perception, and our semantic alignment constraint. The overall loss function is defined as:
\begin{equation}
    \mathcal{L}_{\text{total}} = \mathcal{L}_{\text{rec}} + \mathcal{L}_{\text{adv}} + \lambda_{\text{repa}} \mathcal{L}_{\text{REPA}} \left( + \mathcal{L}_{\text{vq}} \right)
\end{equation}
where the reconstruction loss $\mathcal{L}_{\text{rec}}$ comprises L1 and LPIPS perceptual losses to ensure basic pixel correctness. The generative adversarial loss $\mathcal{L}_{\text{adv}}$ (GAN loss) is retained and specifically dedicated to recovering high-frequency visual textures. Meanwhile, $\mathcal{L}_{\text{REPA}}$ injects macroscopic structure and physical semantics. Finally, $\mathcal{L}_{\text{vq}}$ is included exclusively when training the discrete VideoRAE, accounting for codebook commitment losses.

\begin{figure*}[!htb]
    \centering
    \includegraphics[width=1.0\linewidth]{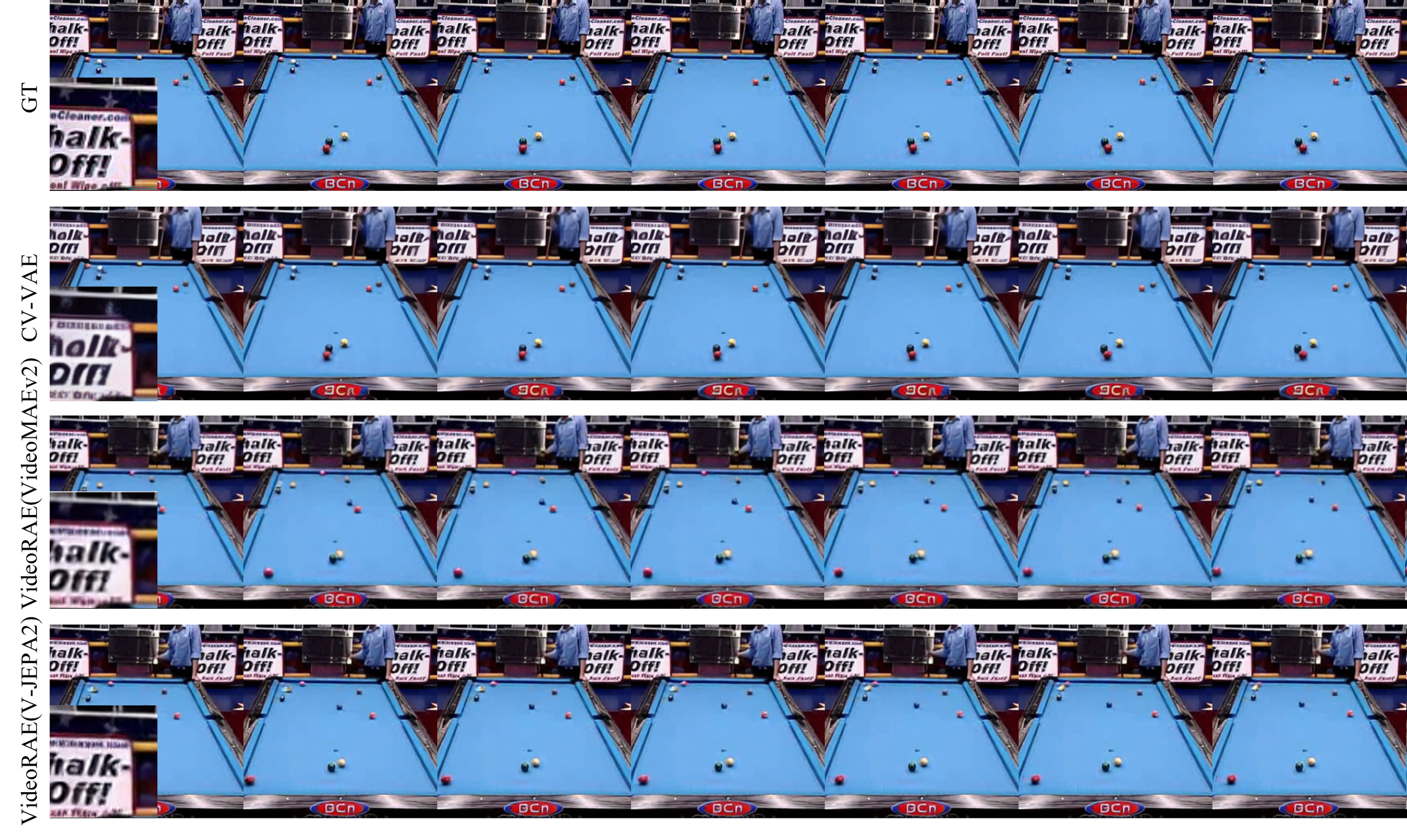}
    \vspace{-5mm}
    \caption{Visualization comparison of the Continuous-space reconstruction on the UCF-101 dataset.}
    \label{fig:vis1}
    \vspace{-13pt}
\end{figure*}

\section{Experiments}

\noindent\textbf{Datasets:} Building upon previous works, we train the VideoRAE on the UCF-101~\citep{soomro2012ucf101} and Kinetics-600 datasets~\citep{carreira218k600}, using 16×256×256 video clips for both training and evaluation; meanwhile, the text-to-video (T2V) generation models are trained specifically on the VideoUFO dataset~\citep{wang2026videoufo}. For the assessment of reconstruction performance, evaluations are conducted on the UCF-101~\citep{soomro2012ucf101} and TokenBench~\citep{agarwal2025cosmos} datasets. Furthermore, we evaluate class-to-video (C2V) generation capabilities on the UCF-101 dataset~\citep{soomro2012ucf101}, and assess text-to-video generation performance on VBench~\citep{huang2024vbench}. Following established evaluation protocols~\citep{wang2024larp}, we sample 10,000 cases to assess the C2V generation of discrete tokenizers, whereas for continuous VAE models, we evaluate using a subset of 2,048 sampled cases.

\noindent\textbf{Implementation details:} For feature extraction, we employ pre-trained and frozen V-JEPA 2~\citep{assran2025vjepa2},V-JEPA 2.1~\citep{murlabadia2026vjepa2_1} and VideoMAEv2~\citep{wang2023videomae} as the teacher models, extracting multi-level intermediate features. The pre-trained video foundation models utilize a $16\times$ spatial downsampling and a $2\times$ temporal downsampling. Subsequently, our 1D self-attention projector compresses the input 3D features into a highly concentrated sequence of 512 or 1024 latent tokens. For the continuous latent space, the channel dimension is linearly projected to a 32/64-dimensional bottleneck. For the discrete latent space, we adopt the Multi-Codebook SimVQ strategy, setting the number of sub-codebooks to $K=4$, with a vocabulary size of $V=4096$ for each sub-codebook. To evaluate the generative performance of the structured latent spaces, we train standard Diffusion Transformers (DiTs)~\citep{DiT} on the continuous VideoRAE, and LLaMA-style Transformer-based autoregressive (AR) models~\citep{llamagen} on the discrete VideoRAE.We also provide experimental results for the 3D variant of the continuous VideoRAE, with a latent shape of 1024 (16*16*4) * 64. All models are trained on NVIDIA H100 GPUs.

\noindent\textbf{Baselines.} 
To comprehensively evaluate the performance of VideoRAE across diverse tasks and architectures, we select appropriate state-of-the-art (SOTA) baselines tailored to each specific experimental setting. Notably, to guarantee fairness, we reproduced and re-tested all open-source baseline models under identical evaluation protocols and environments:

\begin{itemize}[leftmargin=*, topsep=0pt, itemsep=2pt]
    \item \textit{Video Reconstruction.} Under the discrete representation paradigm, we compare our method against leading video tokenizers, including TATS~\citep{ge2022tats}, OmniTokenizer~\citep{wang2024omnitokenizer}, LARP~\citep{wang2024larp}, and SweetTok~\citep{tan2025sweettok}. For continuous autoencoding, we benchmark against not only CV-VAE~\citep{zhao2024cv} and LeanVAE~\citep{cheng2025leanvae}, but also the most recent powerful autoencoders such as LTX-VAE~\citep{hacohen2024ltx}, LTX2~\citep{hacohen2026ltx}, CogVideoX-VAE~\citep{yang2025cogvideox}, and WAN2.1-VAE~\citep{wan2025wan}.
    
    \item \textit{Class-Conditional Generation.} On the UCF-101 dataset, we carefully align the baselines with the downstream generative frameworks. For Autoregressive (AR) models operating in the discrete space, we evaluate against CogVideo~\citep{hong2022cogvideo}, TATS~\citep{ge2022tats}, Video-LaVIT~\citep{jin2024videolavit}, OmniTokenizer~\citep{wang2024omnitokenizer}, LARP~\citep{wang2024larp}, and SweetTok~\citep{tan2025sweettok}. For Diffusion Transformers (DiT) in the continuous space, our comparisons encompass Latte~\citep{ma2024latte}, WF-VAE~\citep{li2025wf}, DeCo-VAE~\citep{yin2025deco}, as well as generative pipelines built upon LeanVAE~\citep{cheng2025leanvae}, LTX-VAE~\citep{hacohen2024ltx}, CogVideoX-VAE~\citep{yang2025cogvideox}, and WAN2.1-VAE~\citep{wan2025wan}.
    
    \item \textit{Text-to-Video Generation.} Finally, for the complex open-domain T2V generation task evaluated on the VBench suite, we compare our continuous VideoRAE directly with the state-of-the-art open-source continuous model, LTX-VAE~\citep{hacohen2024ltx}.
\end{itemize}

\begin{table*}[!htb]
\small
\centering
\setlength{\tabcolsep}{6pt}
\renewcommand{\arraystretch}{1.2}
\caption{Reconstruction performance of discrete tokenizers on UCF-101 and TokenBench.}
\vspace{-2mm}
\begin{tabular}{l c ccc ccc}
\toprule
\multirow{2}{*}{\textbf{Method}} & \multirow{2}{*}{\textbf{\#rToken}} & \multicolumn{3}{c}{\textbf{UCF-101}} & \multicolumn{3}{c}{\textbf{TokenBench-256$\times$256}} \\
\cmidrule(lr){3-5} \cmidrule(lr){6-8}
& & \textbf{PSNR$\uparrow$} & \textbf{LPIPS$\downarrow$} & \textbf{rFVD$\downarrow$} & \textbf{PSNR$\uparrow$} & \textbf{LPIPS$\downarrow$} & \textbf{rFVD$\downarrow$} \\
\midrule

\multicolumn{8}{l}{\textit{AR generative models with discrete video tokenizers}} \\
\hdashline[1pt/2pt]
TATS                     & 1024 & - & - & 162 & - & - & - \\
OmniTokenizer  & 5120 & 29.25 & 0.11 & 28  & 25.55 & 0.11 & 44 \\
LARP-L-Long           & 1024 & 27.88 & 0.12 & 35 & 28.65 & 0.11 & 45 \\
SweetTok                                   & 1280 & 29.27 & \textbf{0.07} & 20 & - & - & - \\

\rowcolor{rowgreen} VideoRAE(VideoMAEv2)    & 1024  & \textbf{29.94} & \underline{0.10} & \underline{16} & \textbf{30.69} & \textbf{0.09} & \underline{33} \\
\rowcolor{rowblue} VideoRAE(V-JEPA 2)         & 1024  & \underline{29.39} & \underline{0.10} & \textbf{13} & \underline{29.93} & \underline{0.10} & \textbf{28} \\

\bottomrule
\end{tabular}
\label{tab:mainresult} 
\end{table*}

\begin{table*}[!htb]
\small
\centering
\setlength{\tabcolsep}{5pt}
\renewcommand{\arraystretch}{1.2}
\caption{Reconstruction performance of continuous autoencoders on UCF-101 and TokenBench.}
\vspace{-2mm}
\begin{tabular}{l c ccc ccc}
\toprule
\multirow{2}{*}{\textbf{Method}} & \multirow{2}{*}{\textbf{Config}} & \multicolumn{3}{c}{\textbf{UCF-101}} & \multicolumn{3}{c}{\textbf{TokenBench-256$\times$256}} \\
\cmidrule(lr){3-5} \cmidrule(lr){6-8}
& & \textbf{PSNR$\uparrow$} & \textbf{LPIPS$\downarrow$} & \textbf{rFVD$\downarrow$} & \textbf{PSNR$\uparrow$} & \textbf{LPIPS$\downarrow$} & \textbf{rFVD$\downarrow$} \\
\midrule

\multicolumn{8}{l}{\textit{Diffusion generative models with continuous video tokenizers}} \\
\hdashline[1pt/2pt]

CV-VAE            & 4096$\times$4 & 30.11 & 0.12 & 57 & 30.80 & 0.11 & 61 \\
LTX-VAE         & 128$\times$128  & \textbf{32.02} & 0.10 & 35 & \textbf{33.74} & \textbf{0.08} & 32 \\
LTX2              & 128$\times$128  & 28.88 & 0.12 & 15 & 29.36 & 0.11 & 36 \\
LeanVAE          & 4096$\times$4 & 30.99 & 0.11 & \textbf{10} & 31.79 & 0.10 & \textbf{23} \\

\rowcolor{rowgreen} VideoRAE(VideoMAEv2)   & 512$\times$32  & \underline{31.23} & \textbf{0.08} & 13 & \underline{32.06} & \textbf{0.08} & \underline{25} \\
\rowcolor{rowblue} VideoRAE(V-JEPA 2)     & 512$\times$32  & 30.40 & \underline{0.09} & \underline{14} & 31.19 & \underline{0.09} & \underline{25} \\
\midrule
CogvideoX-VAE &4096$\times$16 &\textbf{36.22}  &\textbf{0.05}  &7  &\underline{36.35}  &\textbf{0.04}  &\underline{9}\\

WAN2.1-VAE &4096$\times$16 &35.70  &\textbf{0.05} &\textbf{4} &\textbf{37.02} &\textbf{0.04} &\textbf{6}\\

\rowcolor{rowgreen} VideoRAE(VideoMAEv2) &1024$\times$64  &33.64  &\underline{0.06}  &\underline{5} &34.25 &\underline{0.07} &11\\
\rowcolor{rowblue} VideoRAE(V-JEPA 2) &1024$\times$64   &32.14  &0.07  &7  &32.56  &0.08  &13 \\


\bottomrule
\end{tabular}
\label{tab:reconstruction_result} 
\end{table*}

\begin{table*}[!htb]
\small
\centering
\setlength{\tabcolsep}{5pt}
\renewcommand{\arraystretch}{1.2}
\caption{Reconstruction performance of 3D projector variants on UCF-101 and TokenBench.}
\vspace{-2mm}
\begin{tabular}{l c ccc ccc}
\toprule
\multirow{2}{*}{\textbf{Method}} & \multirow{2}{*}{\textbf{Config}} & \multicolumn{3}{c}{\textbf{UCF-101}} & \multicolumn{3}{c}{\textbf{TokenBench-256$\times$256}} \\
\cmidrule(lr){3-5} \cmidrule(lr){6-8}
& & \textbf{PSNR$\uparrow$} & \textbf{LPIPS$\downarrow$} & \textbf{rFVD$\downarrow$} & \textbf{PSNR$\uparrow$} & \textbf{LPIPS$\downarrow$} & \textbf{rFVD$\downarrow$} \\
\midrule

\multicolumn{8}{l}{\textit{Diffusion generative models with continuous video tokenizers}} \\
\hdashline[1pt/2pt]

\rowcolor{rowgreen} VideoRAE-3D(VideoMAEv2) &1024$\times$64  &33.50  &\underline{0.06}  &\underline{5} &34.07 &\underline{0.07} &12\\
\rowcolor{rowblue} VideoRAE-3D(V-JEPA 2) &1024$\times$64   &31.91  &0.07  &7  &32.48  &0.08  &15 \\

\rowcolor{rowyellow} VideoRAE-3D(V-JEPA 2.1) &1024$\times$64   &31.88  &0.07  &7  &32.19  &\underline{0.07}  &16 \\

\bottomrule
\end{tabular}
\label{tab:reconstruction_result_3d} 
\end{table*}

\subsection{Main Results}
\noindent\textbf{Reconstruction Quality Comparison.} 
We comprehensively evaluate the reconstruction capabilities of VideoRAE under both discrete and continuous latent paradigms on the UCF-101\citep{soomro2012ucf101} and TokenBench~\citep{agarwal2025cosmos} datasets. As delineated in Table~\ref{tab:mainresult}, under the discrete tokenizer configuration, VideoRAE establishes new state-of-the-art benchmarks. Specifically, VideoRAE(V-JEPA 2) achieves an excellent rFVD of 13 on UCF-101\citep{soomro2012ucf101} and 28 on TokenBench~\citep{agarwal2025cosmos}, significantly outperforming recent strong baselines such as LARP\citep{wang2024larp} and SweetTok\citep{tan2025sweettok}. Meanwhile, VideoRAE(VideoMAEv2) demonstrates superior spatial fidelity, securing the highest PSNR across both datasets. This verifies the superiority of high-dimensional semantic representations derived from video foundation models. Moreover, it demonstrates that our Multi-Codebook SimVQ strategy can successfully reconstruct high-fidelity pixels even within a highly compressed discrete semantic space.

For the continuous latent space presented in Table~\ref{tab:reconstruction_result}, VideoRAE also delivers highly competitive performance, exhibiting a strong balance between spatial texture recovery and perceptual quality. We observe that while models like LTX-VAE\citep{hacohen2024ltx} achieve high PSNR, they struggle with spatio-temporal consistency (resulting in a sub-optimal rFVD of 35). Conversely, although LeanVAE\citep{cheng2025leanvae} achieves a low rFVD, it relies heavily on a massive latent resolution. In contrast, using a highly compact $512\times 32$ bottleneck, VideoRAE delivers top-tier perceptual quality alongside competitive PSNR performance. Even in comparison with industry-grade VAEs, VideoRAE demonstrates competitive performance despite being trained solely on the K600\citep{carreira218k600} and UCF-101\citep{soomro2012ucf101} datasets. Furthermore, as detailed in Table~\ref{tab:reconstruction_result_3d}, the 3D projector variants of VideoRAE yield similarly strong reconstruction metrics, demonstrating that our VFM-based compression is robust and effective across different architectural choices. The aforementioned reconstruction results across both discrete and continuous models effectively validate our core hypothesis: representations from frozen video foundation models do not hinder reconstruction. Instead, they can surpass traditional pixel-driven autoencoders in reconstructing videos with exceptional perceptual quality and fidelity.

\begin{figure*}[t]
    \vspace{-1mm}
    \centering
    \includegraphics[width=1.0\linewidth]{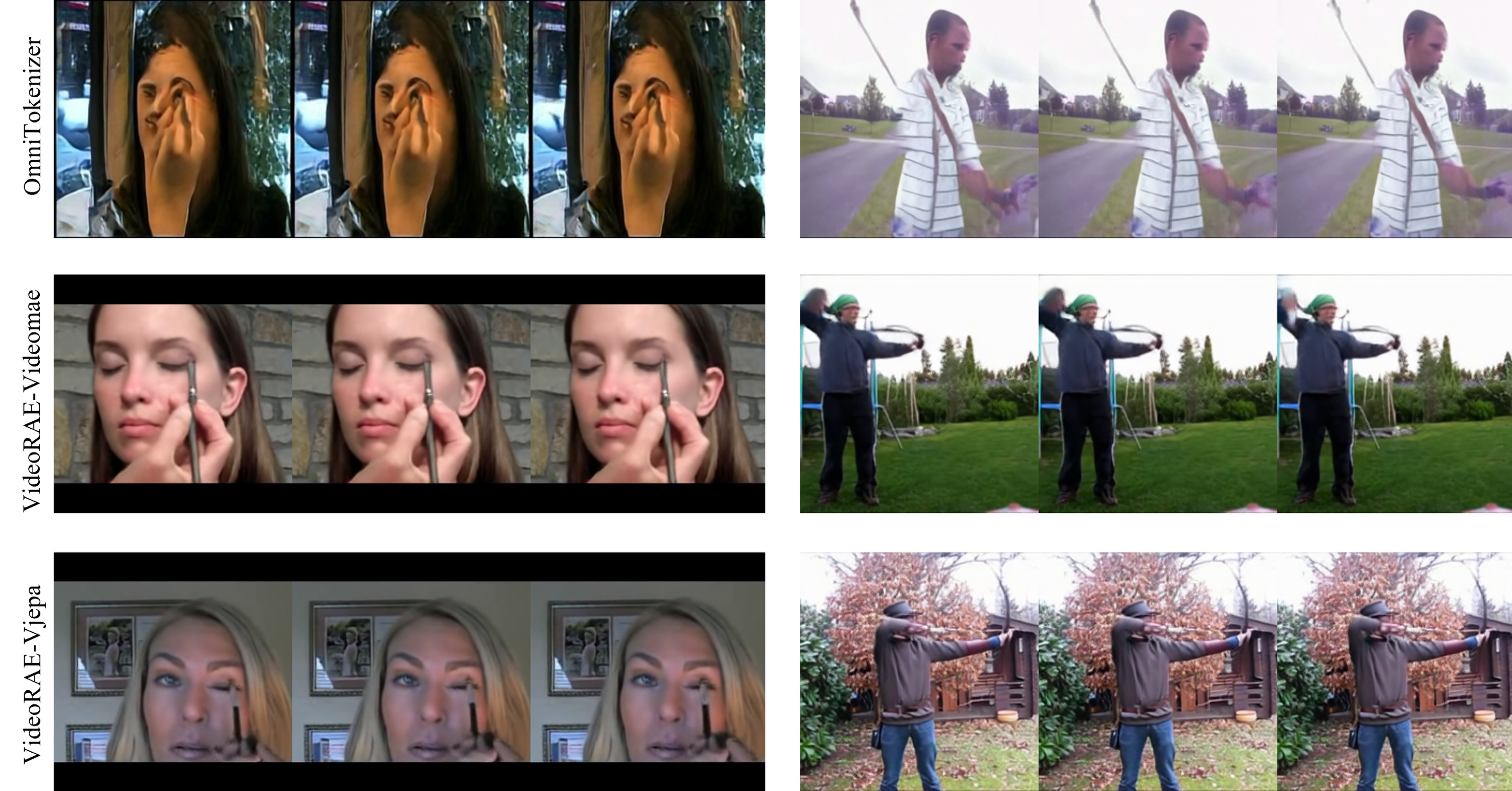}
    \vspace{-5mm}
    \caption{Visualization of the Discrete-token generation results on the UCF-101 dataset.}
    \label{fig:genvis2}
    \vspace{-5pt}
\end{figure*}
\noindent\textbf{Discrete-token class-conditional generation.} 
To evaluate the generative capability of the discrete representations, we train autoregressive (AR) models on our tokenized latent space. As shown in Table~\ref{tab:gen_discrete}, VideoRAE demonstrates highly competitive performance on the UCF-101 class-conditional generation task. Notably, VideoRAE(V-JEPA 2) achieves a state-of-the-art gFVD of 40. Compared to the previously leading method, SweetTok\citep{tan2025sweettok} (which obtains a gFVD of 65 using 1.9B parameters and 1280 tokens), VideoRAE secures a decisive advantage while utilizing fewer tokens and a smaller generator size. The feature space derived from visual foundation models is inherently suited for generative tasks. Furthermore, the discrete tokens obtained through our Multi-Codebook SimVQ quantization and semantic alignment strategy effectively preserve the rich semantics of the VFM, making the downstream autoregressive prediction tasks more efficient and accurate.

\begin{figure*}[t]
    \centering
    \includegraphics[width=1.0\linewidth]{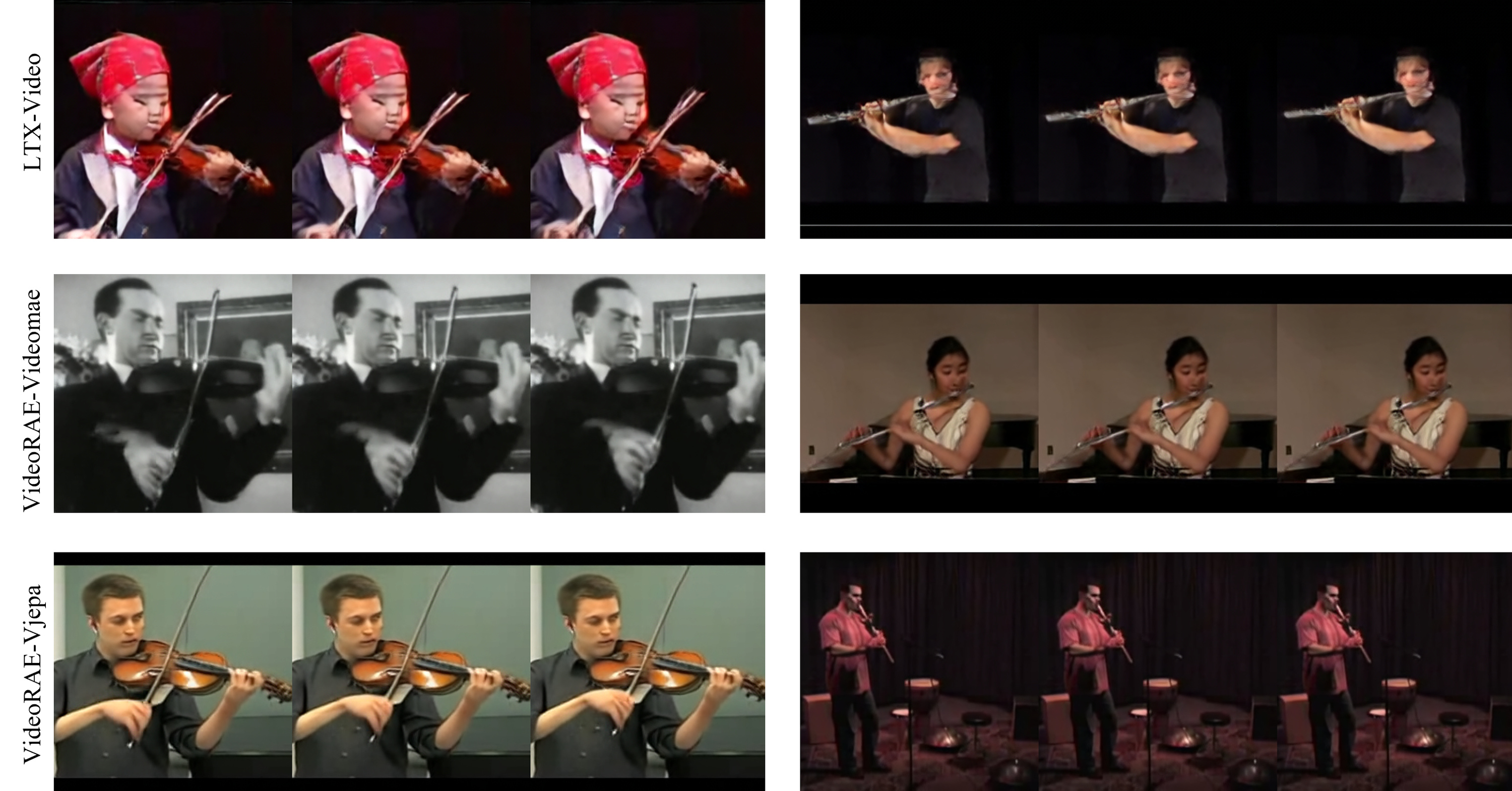}
    \vspace{-5mm}
    \caption{Visualization of the Continuous-space generation results on the UCF-101 dataset.}
    \label{fig:genvis1}
    \vspace{-13pt}
\end{figure*}

\noindent\textbf{Continuous-space class-conditional generation.} 
For the continuous latent space, we adopt a Diffusion Transformer (DiT)\citep{ldm} architecture to evaluate the generation quality. As presented in Table~\ref{tab:gen_continuous}, VideoRAE exhibits strong generative capabilities. While recent advanced continuous VAEs (such as LeanVAE\citep{cheng2025leanvae} and LTX-VAE\citep{hacohen2024ltx}) have improved the baseline gFVD to 164 and 161, respectively, VideoRAE further reduces this metric to 99 (VideoMAEv2) and 93 (V-JEPA 2). Furthermore, we extended industry-grade models, including WAN2.1-VAE\citep{wan2025wan} and CogVideoX\citep{yang2025cogvideox}, to class-to-video generation on the UCF-101 dataset. Although they outperform existing methods like Deco-VAE\citep{yin2025deco}, VideoRAE and its 3D projector variants still deliver a distinct superiority in generation performance, even in comparison with them. This substantial improvement indicates that the KL-free and semantic-aligned continuous latent space provided by VideoRAE offers a structurally sound and generation-friendly manifold for diffusion models, thereby facilitating the synthesis of high-fidelity videos.

\noindent\textbf{Discussion: Reconstruction Fidelity vs. Generative Quality.} 
Across both discrete and continuous modalities, we observe a thought-provoking phenomenon: VideoRAE equipped with VideoMAEv2 consistently exhibits superior pixel-level reconstruction fidelity (i.e., higher PSNR and lower LPIPS). However, our empirical results reveal that ultimate fine-grained reconstruction does not strictly translate to higher generation quality. Conversely, VideoRAE(V-JEPA 2) consistently achieves better gFVD scores across all generative tasks. We attribute this discrepancy to the fundamentally distinct pre-training paradigms of the two video foundation models. VideoMAEv2 is driven by masked pixel reconstruction, which forces the model to heavily focus on fine-grained local motion details. In contrast, V-JEPA 2 utilizes a latent-space prediction mechanism, an objective that compels its representations to be more aligned with high-level semantics and macroscopic spatio-temporal dynamics. Although this deep semantic abstraction slightly compromises absolute pixel-level reconstruction precision, it provides a significantly more generation-friendly latent space that eases the modeling burden on downstream generators. This leads to a crucial conclusion: for achieving high-quality video generation, endowing the latent space with stronger semantic understanding and visual perception is substantially more beneficial than pursuing rigid low-level pixel fidelity.

\begin{table*}[!htb]
\small
\centering
\renewcommand{\arraystretch}{1.2}
\caption{Class-conditional generation evaluation (Discrete AR) on UCF-101, evaluated on 10K generated samples.}
\vspace{-2mm}
\begin{tabularx}{\linewidth}{ l *{4}{>{\centering\arraybackslash}X} }
\toprule
\textbf{Tokenizer} & \textbf{Resolution} &  \textbf{\#Tokens} & \textbf{\#Params} & \textbf{gFVD} $\downarrow$ \\
\midrule
CogVideo~\citep{hong2022cogvideo}       & 256$\times$256 & 6800 & 9.4B & 626 \\
TATS~\citep{ge2022tats}          & 256$\times$256 & 4096 & 321M & 332 \\
Video-LaVIT~\citep{jin2024videolavit}    & 256$\times$256 & 512  & 7B   & 280 \\
OmniTokenizer~\citep{wang2024omnitokenizer}  & 256$\times$256 & 5120 & 650M & 191 \\
LARP-L~\citep{wang2024larp}         & 256$\times$256 & 1024 & 632M & 99  \\
SweetTok~\citep{tan2025sweettok}       & 256$\times$256 & 1280 & 650M & 84  \\
SweetTok~\citep{tan2025sweettok}       & 256$\times$256 & 1280 & 1.9B & 65  \\
\midrule
\rowcolor{rowgreen} VideoRAE(VideoMAEv2) & 224$\times$224 & 1024 & 1.3B & \underline{45} \\
\rowcolor{rowblue} VideoRAE(V-JEPA 2)   & 256$\times$256 & 1024 & 1.3B & \textbf{40} \\
\bottomrule
\end{tabularx}
\label{tab:gen_discrete}
\end{table*}

\begin{table*}[!htb]
\small
\centering
\renewcommand{\arraystretch}{1.2}
\caption{Class-conditional generation evaluation (Continuous DiT) on UCF-101, evaluated on 2K generated samples.}
\vspace{-2mm}
\begin{tabularx}{\linewidth}{ l *{3}{>{\centering\arraybackslash}X} }
\toprule
\textbf{Method} & \textbf{Resolution} & \textbf{Configs} & \textbf{gFVD} $\downarrow$ \\
\midrule

Latte~\citep{ma2024latte}          & 256$\times$256 & 4096*4  & 478 \\
LeanVAE~\citep{cheng2025leanvae}        & 256$\times$256 & 4096*4  & 164 \\
LTX-VAE~\citep{hacohen2024ltx}        & 256$\times$256 & 128*128 & 161 \\ 
\rowcolor{rowgreen} VideoRAE(VideoMAEv2) & 224$\times$224 & 512*32  & \underline{99}  \\
\rowcolor{rowblue} VideoRAE(V-JEPA 2)   & 256$\times$256 & 512*32  & \textbf{93}  \\

\midrule 

WF-VAE~\citep{li2025wf}         & 256$\times$256 & 4096*16 & 371 \\
LeanVAE~\citep{cheng2025leanvae}          & 256$\times$256 & 4096*16 & 175 \\
DeCo-VAE ~\citep{yin2025deco}      & 256$\times$256 & 4096*16 & 166 \\ 
CogvideoX-VAE~\citep{yang2025cogvideox}  & 256$\times$256 & 4096*16 & 151 \\
WAN2.1-VAE~\citep{wan2025wan}     & 256$\times$256 & 4096*16 & 126 \\
\rowcolor{rowgreen} VideoRAE(VideoMAEv2) & 224$\times$224 & 1024*64 & \underline{103} \\
\rowcolor{rowblue} VideoRAE(V-JEPA 2)   & 256$\times$256 & 1024*64 & \textbf{94}  \\
\bottomrule
\end{tabularx}
\label{tab:gen_continuous}
\end{table*}

\begin{table*}[!htb]
\small
\centering
\renewcommand{\arraystretch}{1.2}
\caption{Class-conditional generation evaluation (3D projector variants) on UCF-101, evaluated on 2K generated samples.}
\vspace{-2mm}
\begin{tabularx}{\linewidth}{ l *{3}{>{\centering\arraybackslash}X} }
\toprule
\textbf{Method} & \textbf{Resolution} & \textbf{Configs} & \textbf{gFVD} $\downarrow$ \\
\midrule
\rowcolor{rowgreen} VideoRAE-3D(VideoMAEv2) & 224$\times$224 & 1024*64 & 104 \\
\rowcolor{rowblue} VideoRAE-3D(V-JEPA 2)   & 256$\times$256 & 1024*64 & \underline{96}  \\
\rowcolor{rowyellow} VideoRAE-3D(V-JEPA 2.1)   & 256$\times$256 & 1024*64 & \textbf{93}  \\
\bottomrule
\end{tabularx}
\label{tab:gen_continuous_3d}
\end{table*}

\begin{figure}[!htb]
    \centering
    \includegraphics[width=0.49\linewidth]{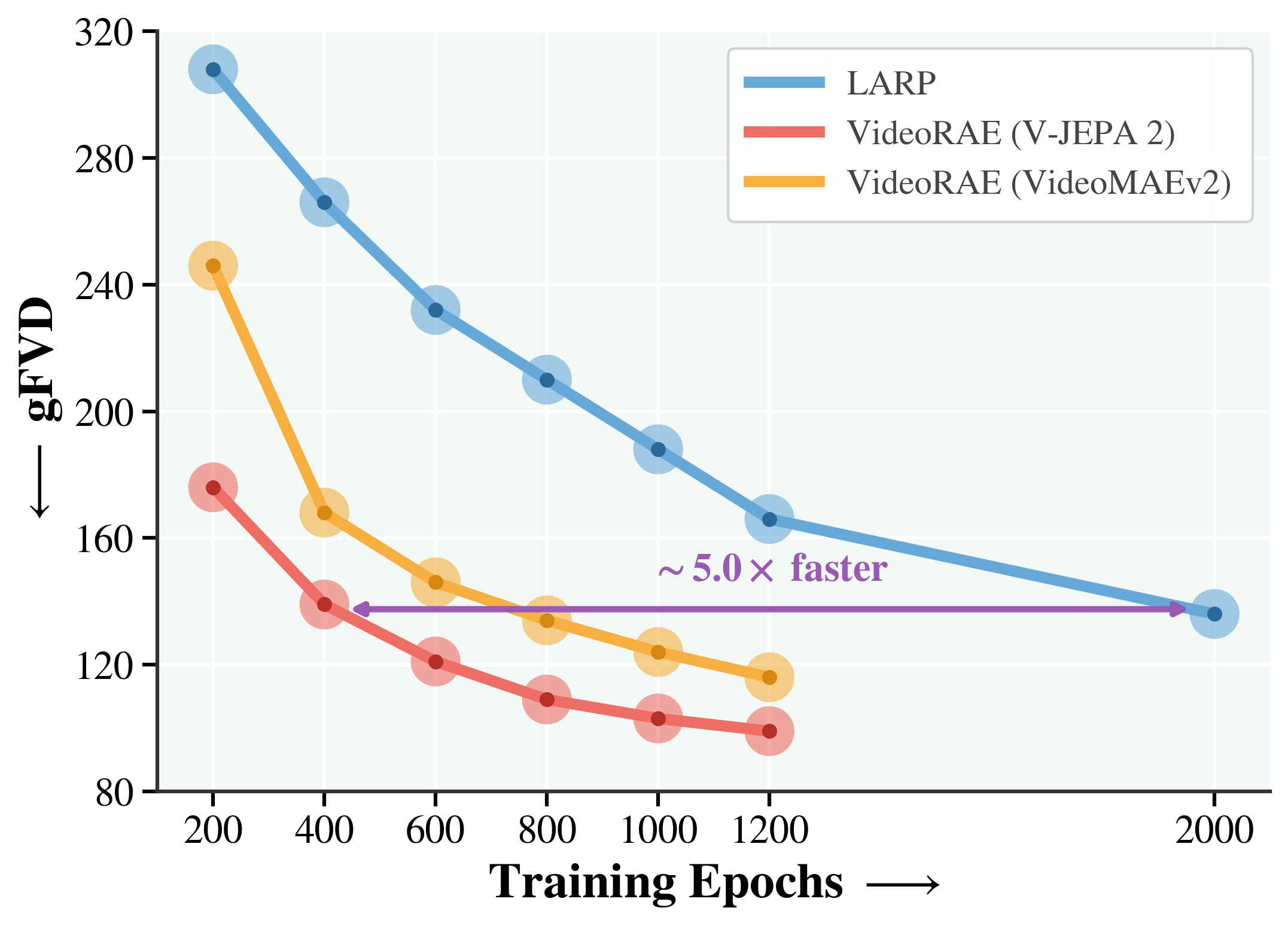}\hfill
    \includegraphics[width=0.49\linewidth]{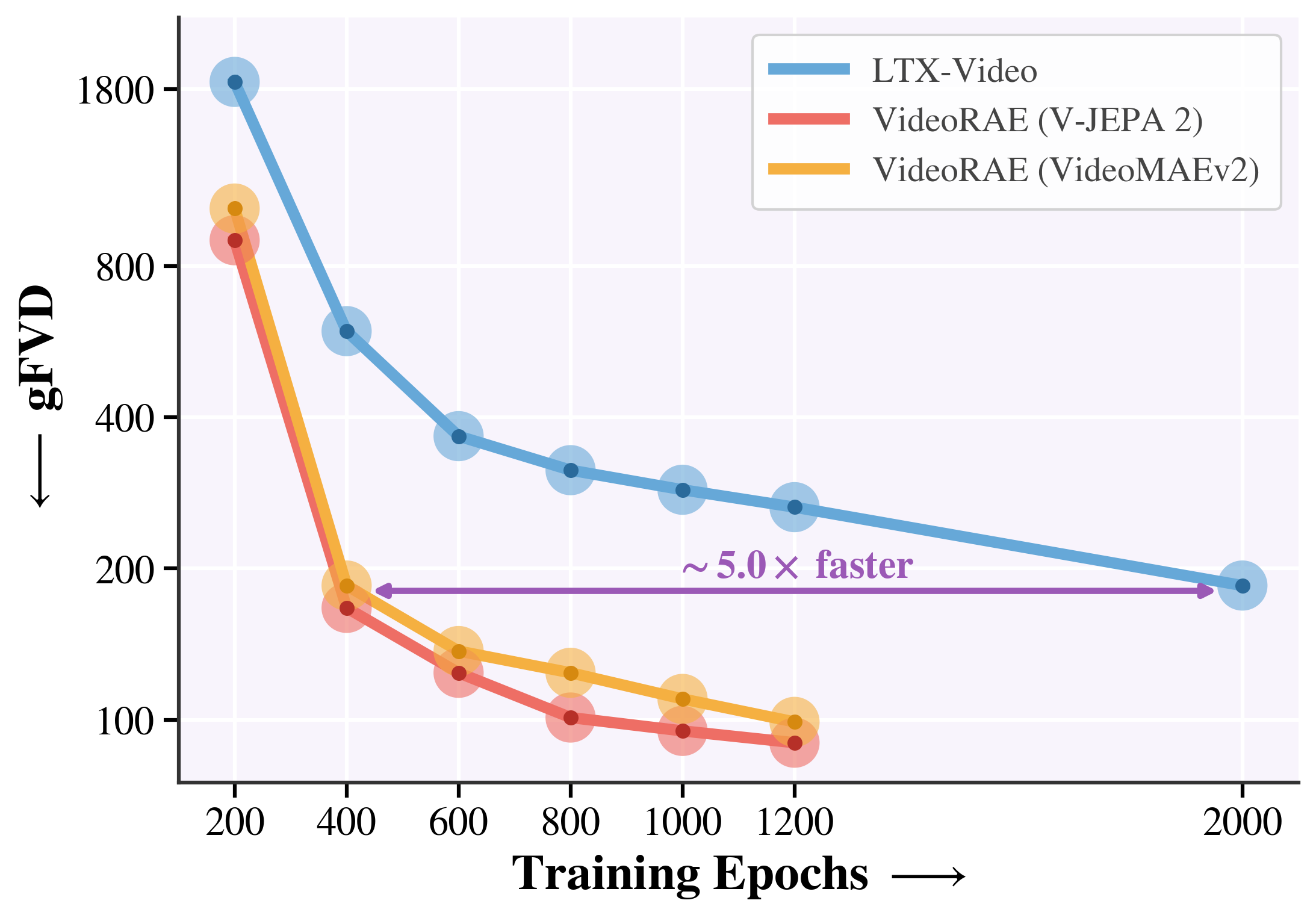} 
    
    \caption{Training convergence speed comparison in AR (left) and DiT (right) generative modeling.}
    \label{fig:gfidspeed}
\end{figure}

\noindent\textbf{Accelerated Convergence in Generative Training.} 
To further illustrate the efficiency of our semantic-driven latent space, we first compare the generative training convergence speed between VideoRAE and the strong baseline LARP under the discrete tokenizer setting, with 2,000 generated samples evaluated for each model. As illustrated in Figure~\ref{fig:gfidspeed}, the autoregressive generative model trained on the traditional pixel-driven LARP tokenizer requires $2000$ epochs to achieve a gFVD score of $136$. In stark contrast, the model trained on VideoRAE's representations reaches a comparable gFVD of $139$ at merely $400$ epochs, demonstrating an impressive $\sim 5.0\times$ acceleration in training convergence. Similarly, within the continuous latent space, VideoRAE exhibits a comparable trend of significantly accelerating model convergence when evaluated against LTX-VAE. By alleviating the immense burden on the generator of learning complex spatio-temporal dynamics from scratch, VideoRAE enables downstream generative models to focus entirely on learning high-level sequence transition logic. This fundamentally validates our claim that a semantically aligned latent space is pivotal for reducing generative training costs.

\begin{figure}[!htb]
    \centering
    \begin{subfigure}{0.33\linewidth}
        \centering
        \includegraphics[width=\linewidth]{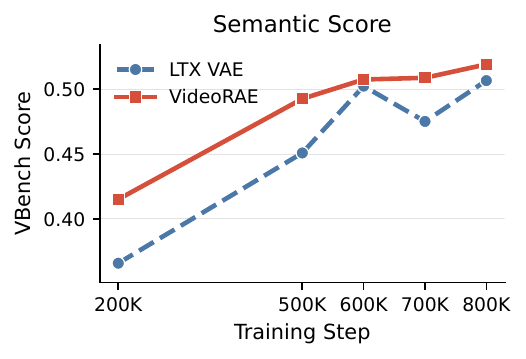} 
        \label{fig:vbench_semantic}
    \end{subfigure}\hfill
    \begin{subfigure}{0.33\linewidth}
        \centering
        \includegraphics[width=\linewidth]{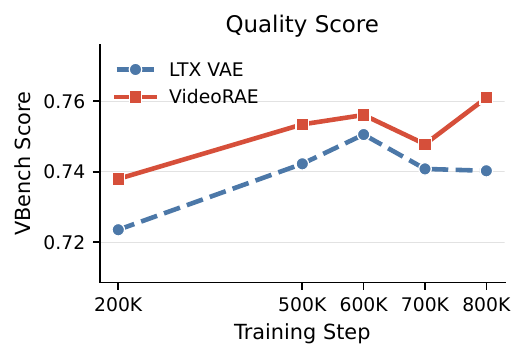} 
        \label{fig:vbench_quality}
    \end{subfigure}\hfill
    \begin{subfigure}{0.33\linewidth}
        \centering
        \includegraphics[width=\linewidth]{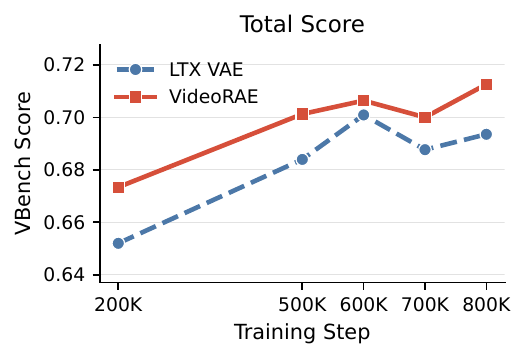} 
        \label{fig:vbench_total}
    \end{subfigure}
    \vspace{-4mm}
    \caption{Convergence comparison between VideoRAE and LTX-VAE on VBench.}
    \label{fig:vbench_convergence}
\end{figure}

\begin{table}[ht]
\centering
\caption{Text-to-video generation evaluation on VBench. VideoRAE achieves stronger overall performance than LTX-VAE, with clear gains in both visual consistency and semantic alignment.}
\label{tab:vbench}
\resizebox{\linewidth}{!}{
\begin{tabular}{lccccccccc}
\toprule
\textbf{Method}
& \textbf{Total}
& \textbf{Quality}
& \textbf{Semantic}
& \textbf{Subject}
& \textbf{Background}
& \textbf{Imaging}
& \textbf{Object}
& \textbf{Human}
& \multirow{2}{*}{\textbf{Scene $\uparrow$}} \\
& \textbf{Score $\uparrow$}
& \textbf{Score $\uparrow$}
& \textbf{Score $\uparrow$}
& \textbf{Cons. $\uparrow$}
& \textbf{Cons. $\uparrow$}
& \textbf{Qual. $\uparrow$}
& \textbf{Class $\uparrow$}
& \textbf{Action $\uparrow$}
& \\
\midrule
LTX-VAE
&69.35 &74.03 &50.67 & 84.14 &95.51 &43.54 &58.51 &54.50 &23.29 \\
\rowcolor{rowblue}
VideoRAE (Ours)
& \textbf{71.26} & \textbf{76.09} & \textbf{51.92} & \textbf{91.27} & \textbf{97.02} & \textbf{46.02} & \textbf{59.07} & \textbf{60.80} & \textbf{27.57} \\
\bottomrule
\end{tabular}
}
\end{table}

\noindent\textbf{Text-to-Video Generation on VBench.}  To further evaluate the generative capacity of VideoRAE in complex open-domain scenarios, we extend our study to text-to-video generation. We build our T2V system on top of the OpenSora2~\citep{opensora2} training framework and replace its original video autoencoder with either LTX-VAE~\citep{hacohen2024ltx} or our trained VideoRAE, while keeping the diffusion backbone and training recipe comparable. Specifically, we train a 2B-scale MMDiT-style diffusion transformer, consisting of dual-stream text-video blocks followed by single-stream multimodal blocks, with hidden size 2048, 16 attention heads, 8 dual-stream blocks, and 16 single-stream blocks. The model is conditioned on T5 and CLIP text embeddings and trained to predict the flow-matching target in the latent space of each video tokenizer. We train both systems on VideoUFO~\citep{wang2026videoufo} at 256 × 256 resolution and 12 FPS, using 17-frame clips for LTX-VAE and 16-frame clips for VideoRAE, matching the native temporal interface of each autoencoder. Both models are trained with a global batch size of 256 for 800K optimization steps. 

During training, we periodically evaluate intermediate checkpoints on VBench~\citep{huang2024vbench} and report the evolution of Total Score, Quality Score, and Semantic Score in Figure~\ref{fig:vbench_convergence}. As shown in the figure, the VideoRAE-based generator converges faster than the LTX-VAE counterpart and consistently achieves better scores throughout training. After convergence, we further report detailed VBench metrics in Table~\ref{tab:vbench}.
These results are consistent with our C2V observations and suggest that VideoRAE provides a more effective latent space for generative modeling. By using a visual foundation model as the encoder, VideoRAE produces latents that are both semantically enriched and highly compact, allowing the diffusion transformer to model high-level video content more efficiently. This improved representation reduces the burden on the diffusion model to rediscover semantic structure from low-level reconstruction latents, leading to faster convergence and stronger final generation quality.

Separately, we investigate whether VideoRAE can be integrated into an existing large-scale text-to-video model. We initialize the generator from an 11B OpenSora2 checkpoint, replace its original autoencoder with VideoRAE-3D, and fine-tune the adapted model on 5M video-text training samples. The grid-structured latent interface of VideoRAE-3D facilitates its integration with the pretrained video diffusion backbone. Qualitative samples from this setting are shown in Figure~\ref{fig:videoraet2v}, with additional video samples available on our \href{https://zhxie0117.github.io/VideoRAE/}{project page}. These high-quality generations demonstrate the practical effectiveness of VideoRAE as a latent representation for large-scale text-to-video synthesis, complementing the controlled 2B from-scratch VBench evaluation.

\subsection{Ablation Study}

\noindent\textbf{Quantization Strategies.}
We ablate the impact of four quantization methods on the VideoRAE model: VQ, SimVQ~\citep{zhu2025simvq}, MCQ, and Multi-codebook SimVQ.The traditional Vector Quantization yields the poorest reconstruction and generation performance. We attribute this to codebook collapse and non-convergence, which are caused by the excessively low quantization dimensionality when dealing with highly semantic latent spaces. In contrast, SimVQ scales the quantization dimension to 128, bringing a noticeable improvement. When transitioning to MCQ, it achieves a significant performance boost compared to the single-codebook counterpart. This improvement, in our view, stems from MCQ's ability to effectively expand the codebook capacity while avoiding the optimization difficulties associated with large codebooks. These results provide qualitative evidence that VideoRAE is not only effective in controlled benchmarks, but can also serve as a practical and scalable latent interface for large-scale text-to-video generation.

\begin{table}[!htb]
\small
\renewcommand{\arraystretch}{1.2}
\caption{Ablation study on quantization strategies.}
\vspace{-2mm}
\begin{tabularx}{\linewidth}{ l *{5}{>{\centering\arraybackslash}X} }
\toprule
\textbf{Quantization Method} & \textbf{PSNR$\uparrow$} & \textbf{SSIM$\uparrow$} & \textbf{LPIPS$\downarrow$} & \textbf{rFVD$\downarrow$} & \textbf{gFVD$\downarrow$} \\
\midrule
VQ                   & 25.50 & 0.82 & 0.14 & 35 & 95 \\
SimVQ                & 28.32 & 0.87 & 0.12 & 24 & 59 \\
MCQ                  & 29.10 & 0.90 & 0.10 & 16 & 51 \\
\rowcolor{rowblue} Multi-codebook SimVQ & \textbf{29.39} & \textbf{0.91} & \textbf{0.10} & \textbf{13} & \textbf{40} \\
\bottomrule
\end{tabularx}
\label{tab:ablation_quantization} 
\end{table}

\noindent\textbf{Representation Alignment.} 
To evaluate the effectiveness of our proposed Representation Alignment (REPA) module, we conduct ablation experiments by training both continuous and discrete models with and without the REPA objective. As shown in Table~\ref{tab:ablation_repa}, the integration of REPA consistently yields significant performance improvements across most evaluation metrics in both latent spaces. Specifically, incorporating REPA reduces the gFVD from 105 to 93 in the continuous setting, and from 67 to 40 in the discrete setting, while also substantially improving perceptual metrics. Furthermore, we observe that applying a KL penalty in the continuous setting compromises both reconstruction and generation quality to some extent. This demonstrates that aligning the decoder's features with the VFM teacher at both local and global scales not only improves the model's perceptual quality, but also acts as a strong semantic constraint that successfully regularizes the latent space topology without requiring any KL penalty, rendering this high-quality latent distribution more conducive to downstream generation tasks.



\begin{table}[!htb]
\small
\renewcommand{\arraystretch}{1.2}
\caption{Ablation study on the effect of REPA for semantic alignment.}
\vspace{-2mm}
\begin{tabularx}{\linewidth}{ l *{6}{>{\centering\arraybackslash}X} }
\toprule
\textbf{Latent Space} & \textbf{REPA} & \textbf{PSNR$\uparrow$} & \textbf{SSIM$\uparrow$} & \textbf{LPIPS$\downarrow$} & \textbf{rFVD$\downarrow$} & \textbf{gFVD$\downarrow$} \\
\midrule
Continuous    & w/o & 30.57 & 0.91 & 0.10 & 18 & 105 \\
Continuous-kl & w/o & 27.45 & 0.81 & 0.12 & 31 & 101 \\
\rowcolor{rowblue}
Continuous    & w/  & 30.40 & 0.91 & 0.10 & 14 & 93 \\
\midrule
Discrete      & w/o & 29.71 & 0.90 & 0.10 & 15 & 67 \\
\rowcolor{rowblue}
Discrete      & w/  & 29.39 & 0.91 & 0.10 & 13 & 40 \\
\bottomrule
\end{tabularx}
\label{tab:ablation_repa} 
\end{table}

\noindent\textbf{Multi-Scale Feature Aggregation.} 
We investigate the impact of aggregating features from different levels of the visual foundation model (VFM) teacher. As shown in Table~\ref{tab:ablation_multiscale}, relying solely on deep layers yields sub-optimal reconstruction results. In terms of the perceptual quality metric, rFVD, whether shallow features are incorporated does not lead to a significant discrepancy. However, since deep features primarily capture high-level semantics while discarding fine-grained spatial details, they exhibit an insufficient capacity to restore high-frequency details, resulting in a severely compromised PSNR. By progressively incorporating shallower layers, all reconstruction metrics steadily improve. Ultimately, utilizing multi-level features (Layers 8--24) achieves the best trade-off between reconstruction and generation performance, yielding a PSNR of 29.39 and a gFVD of 40. This highlights the necessity of multi-scale feature aggregation to simultaneously capture both macro-semantics and micro-level texture details.

\begin{table}[!htb]
\small
\renewcommand{\arraystretch}{1.2}
\caption{Ablation study on multi-scale feature aggregation.}
\vspace{-2mm}
\begin{tabularx}{\linewidth}{ l *{5}{>{\centering\arraybackslash}X} }
\toprule
\textbf{Aggregated Layers} & \textbf{PSNR$\uparrow$} & \textbf{SSIM$\uparrow$} & \textbf{LPIPS$\downarrow$} & \textbf{rFVD$\downarrow$} & \textbf{gFVD$\downarrow$} \\
\midrule
Layer 24       & 26.59 & 0.85 & 0.14 & 22 & 51 \\
Layers 16--24  & 28.35 & 0.87 & 0.12 & 18 & 41 \\
Layers 12--24  & 29.11 & 0.90 & 0.11 & 15 & 42 \\
\rowcolor{rowblue} 
Layers 8--24   & \textbf{29.39} & \textbf{0.91} & \textbf{0.10} & \textbf{13} & \textbf{40} \\
\bottomrule
\end{tabularx}
\label{tab:ablation_multiscale} 
\end{table}

\noindent\textbf{Feature Fusion Strategies.} 
We compare four feature fusion strategies for aggregating multi-layer VFM representations: Element-wise Addition, Decoupled Fusion, Gated Fusion, and AdaLN. The results are summarized in Table~\ref{tab:ablation_fusion}. Empirically, the more complex Decoupled Fusion strategy and the simple Element-wise Addition perform on par with each other in terms of both reconstruction and generation quality, both slightly outperforming Gated Fusion and AdaLN. Given the advantages of simplicity, high computational efficiency, and zero additional parameter overhead, we select Element-wise Addition as our default feature fusion strategy.


\begin{table}[!htb]
\small
\renewcommand{\arraystretch}{1.2}
\caption{Ablation study on feature fusion strategies.}
\vspace{-2mm}
\begin{tabularx}{\linewidth}{ l *{5}{>{\centering\arraybackslash}X} }
\toprule
\textbf{Fusion Strategy} & \textbf{PSNR$\uparrow$} & \textbf{SSIM$\uparrow$} & \textbf{LPIPS$\downarrow$} & \textbf{rFVD$\downarrow$} & \textbf{gFVD$\downarrow$} \\
\midrule
Decoupled Fusion             & 29.43 & 0.91 & 0.10 & 13 & 43 \\
Semantic-Guided AdaLN Fusion & 29.03 & 0.90 & 0.11 & 18 & 51 \\
Gated Fusion                 & 29.10 & 0.90 & 0.11 & 16 & 49 \\
\rowcolor{rowblue}
Element-wise Add (Ours)      & 29.39 & 0.91 & 0.10 & 13 & 40 \\
\bottomrule
\end{tabularx}
\label{tab:ablation_fusion} 
\end{table}

\section{Conclusion}
In this paper, we introduced VideoRAE, a semantic-driven video autoencoder that bridges the gap between visual understanding and generation. By challenging the prevailing assumption that frozen Video Foundation Models (VFMs) are unsuitable for pixel-level reconstruction, we demonstrated that VFM representations can serve as highly robust latent spaces for generative modeling. Designed as a unified framework, VideoRAE seamlessly supports both continuous and discrete generative paradigms. Through our Representation Alignment (REPA) objective, the latent manifold is intrinsically regularized, completely obviating the need for traditional distribution constraints. Extensive experiments confirm that VideoRAE not only achieves superior reconstruction quality but significantly accelerates the convergence of downstream generative models. Most importantly, our findings reveal that a semantically structured latent space is far more critical for video generation than rigid pixel-level fidelity. By establishing VFMs as the new default latent foundation, VideoRAE paves a promising pathway toward unifying representation learning and generative modeling.


\bibliographystyle{iclr2026_conference} 
\bibliography{iclr2026_conference} 
\end{document}

%% file: math_commands.tex
\usepackage{amsmath,amsfonts,bm}

\def\eqref#1{equation~\ref{#1}}

\def\1{\bm{1}}

\DeclareMathAlphabet{\mathsfit}{\encodingdefault}{\sfdefault}{m}{sl}
\SetMathAlphabet{\mathsfit}{bold}{\encodingdefault}{\sfdefault}{bx}{n}

